\documentclass{article}

\PassOptionsToPackage{numbers, compress}{natbib}
    \usepackage[preprint]{neurips_2025}

\usepackage[utf8]{inputenc} % allow utf-8 input
\usepackage[T1]{fontenc}    % use 8-bit T1 fonts
\usepackage{hyperref}       % hyperlinks
\usepackage{url}            % simple URL typesetting
\usepackage{booktabs}       % professional-quality tables
\usepackage{amsfonts}       % blackboard math symbols
\usepackage{nicefrac}       % compact symbols for 1/2, etc.
\usepackage{microtype}      % microtypography
\usepackage[table,x11names]{xcolor}       % colors

\usepackage{hyperref}       % hyperlinks
\hypersetup{
colorlinks= true,
urlcolor = blue,
linkcolor = red,
citecolor = blue,
}

\usepackage{xspace}
\usepackage{graphicx}
\usepackage{multicol}
\usepackage{multirow}
\usepackage{makecell}
\usepackage{listings}
\usepackage{xcolor}  % for optional coloring
\newcommand{\datasetname}{\textsc{BioDSA}-1K\xspace}
\newcommand{\err}{\mathcal{E}}

\title{\datasetname: Benchmarking Data Science Agents for Biomedical Research}

\author{%
Zifeng Wang\thanks{Equal contribution.} \quad \textbf{Benjamin Danek}$^{*}$  \quad \textbf{Jimeng Sun} \\
University of Illinois Urbana-Champaign \\
Correspondence: \texttt{\{zifengw2,jimeng\}@illinois.edu} \\
\url{https://ryanwangzf.github.io/projects/biodsa}
}

\PassOptionsToPackage{numbers}{natbib}
\usepackage{wrapfig}
\usepackage{subcaption}
\usepackage{enumitem}
\usepackage{titletoc}
\usepackage{amsmath}
\usepackage{amssymb}
\newcommand\DoToC{%
  \startcontents
  \printcontents{}{1}{\textbf{Contents of Appendix}\vskip3pt\hrule\vskip5pt}
  \vskip3pt\hrule\vskip5pt
}

\begin{document}

\maketitle

\begin{abstract}
Validating scientific hypotheses is a central challenge in biomedical research, and remains difficult for artificial intelligence (AI) agents due to the complexity of real-world data analysis and evidence interpretation. In this work, we present \datasetname{}, a benchmark designed to evaluate AI agents on realistic, data-driven biomedical hypothesis validation tasks. \datasetname{} consists of 1,029 hypothesis-centric tasks paired with 1,177 analysis plans, curated from over 300 published biomedical studies to reflect the structure and reasoning found in authentic research workflows. Each task includes a structured hypothesis derived from the original study's conclusions, expressed in the affirmative to reflect the language of scientific reporting, and one or more pieces of supporting evidence grounded in empirical data tables. While these hypotheses mirror published claims, they remain testable using standard statistical or machine learning methods. The benchmark enables evaluation along four axes: (1) hypothesis decision accuracy, (2) alignment between evidence and conclusion, (3) correctness of the reasoning process, and (4) executability of the AI-generated analysis code. Importantly, \datasetname{} includes non-verifiable hypotheses: cases where the available data are insufficient to support or refute a claim, reflecting a common yet underexplored scenario in real-world science. We propose \datasetname{} as a foundation for building and evaluating generalizable, trustworthy AI agents for biomedical discovery.
\end{abstract}

\section{Introduction}
Artificial intelligence (AI) agents promise to accelerate scientific discovery~\cite{xi2025rise,guo2024large}, with the emergence of ``AI scientists''~\cite{gottweis2025towards} capable of collaborating with human researchers to perform research tasks such as literature mining and data analysis~\cite{boiko2023autonomous,wang2025foundation,majumder2024position,gao2024empowering}. Large language models (LLMs)~\cite{achiam2023gpt} can serve as the intelligence backbone for converting natural language to structured outputs such as code and mathematical expressions. As a core task in biomedical research, data science bridges the gap from proposed hypotheses to novel discoveries leveraging biomedical data. For example, a researcher might hypothesize that ``{Genes involved in histone modification `` are frequently mutated in non-Hodgkin lymphoma.}'' Testing such hypotheses often requires close collaboration between biomedical experts and data scientists to design analyses, write code, and interpret results, and thus far has been mainly manual efforts in practice~\cite{meyer2019healthcare}.

\begin{figure}[ht]
    \centering
    \begin{subfigure}[t]{0.49\textwidth}
        \centering
        \includegraphics[width=\linewidth]{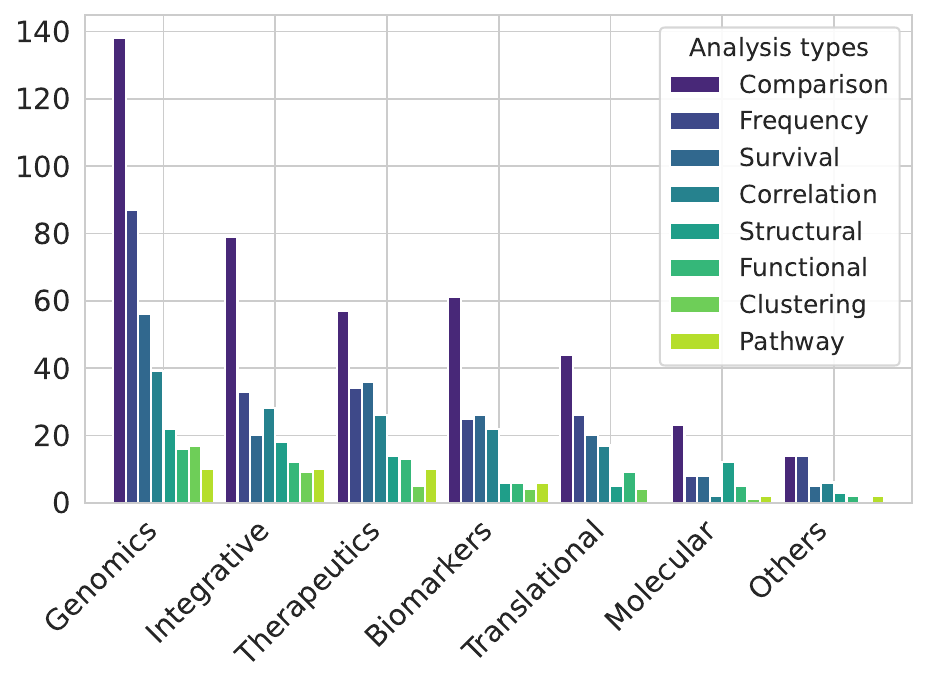}
    \end{subfigure}
    \hfill
    \begin{subfigure}[t]{0.49\textwidth}
        \centering
        \includegraphics[width=\linewidth]{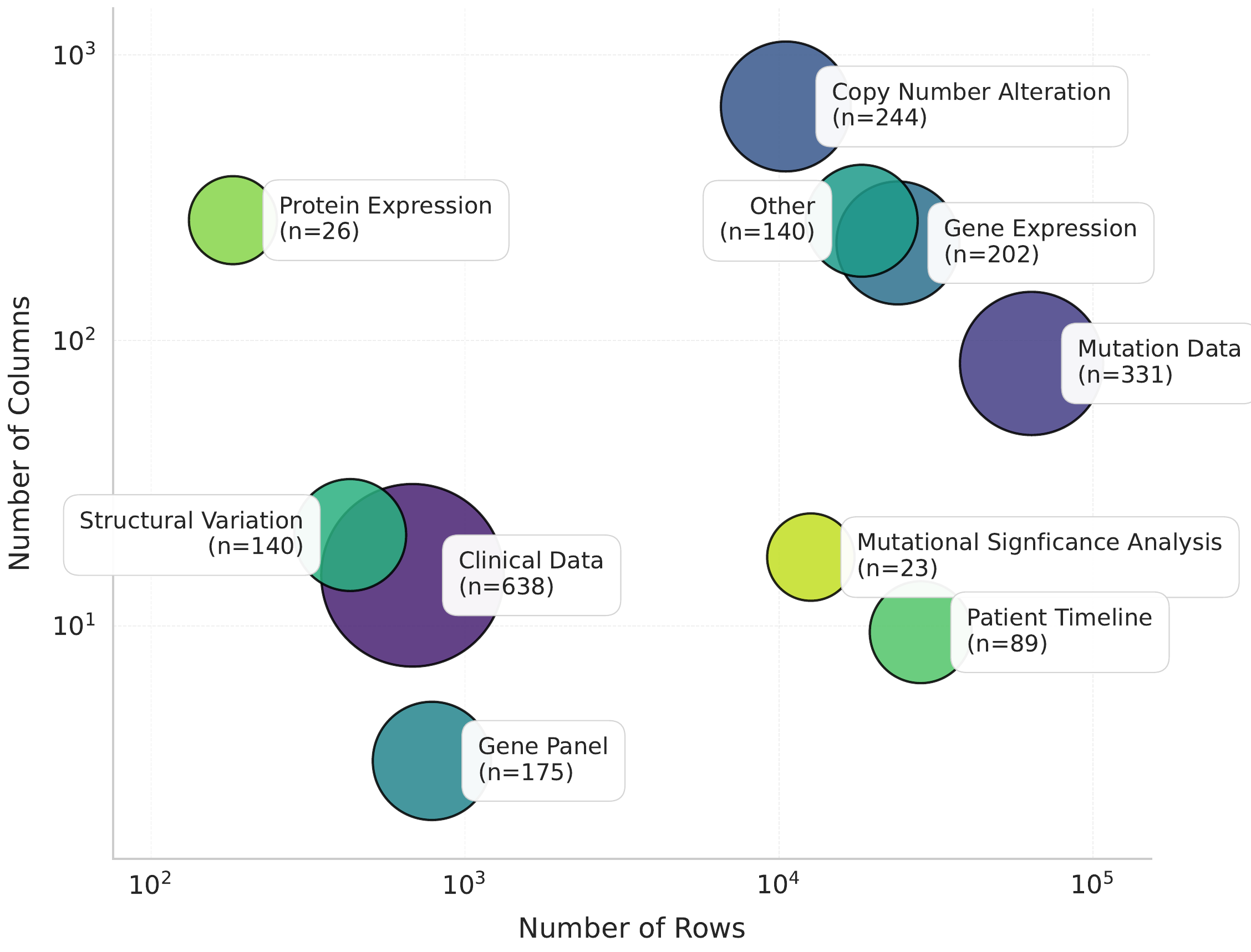}
    \end{subfigure}
    \caption{Benchmark statistics. (left) \datasetname includes diverse types of biomedical research and data analysis tasks created from 329 publications; the x-axis indicates the publication types.; (Right) Bubble plot illustrating the diverse range of biomedical data tables in \datasetname, showing each data table's number of rows (x-axis, log-scale) versus number of columns (y-axis, log-scale).}
    \label{fig:data_stats_combined}
    \vspace{-1em}
\end{figure}

Recent efforts have demonstrated LLM-based agents capable of designing experiments, generating code, and summarizing results~\cite{wang2024can,roohani2024biodiscoveryagent,tang2024biocoder,huang2025automated}. However, existing systems often focus on narrow tasks within biomedical research or are evaluated on limited scenarios. In this work, we aim to systematically investigate the following research question: \textit{To what extent can state-of-the-art LLMs and AI agents perform data science tasks in biomedical research?} Answering this question requires a curated benchmark dataset that captures the breadth and complexity of data science tasks in biomedical research. The following challenges have not been fully explored: (1) although previous studies leverage publications to create data science tasks~\cite{wang2024can,majumderdiscoverybench,chen2024scienceagentbench}, those test cases are drawn from a small number of papers, which may not reflect the full scope of biomedical research; (2) limited task diversity as a consequence of restricted case selection; (3) the involved tasks are performed on relatively simple datasets, such as one or two tables with tens of columns; (4) overlooking the foundational data analysis steps and observed evidence that support or refute a hypothesis, thus, correct hypothesis prediction alone does not guarantee the agent performed the correct analysis; and (5) the inclusion of non-verifiable hypotheses, where the required data is absent or insufficient to support a conclusive answer, yet such cases are rarely discussed.

In this paper, we introduce \datasetname (\textbf{Bio}medical \textbf{D}ata \textbf{S}cience \textbf{A}gent Benchmark), a novel framework for evaluating AI agents on biomedical data science research tasks (Figure 1). \datasetname specifies a complete cycle of hypothesis formulation, data analysis, and validation, by curating detailed experimental components extracted from published biomedical studies. Specifically, each instance includes a hypothesis statement, corresponding analysis plans, evidence summaries, and quantitative outcome measures.

As illustrated in Figure~\ref{fig:data_stats_combined}, \datasetname includes 1,029 scientific hypotheses and the corresponding 1,177 analysis tasks drawn from 329 publications of eight types of publications. The analysis tasks are also comprehensive in terms of common analysis is done in biomedical research.  A comparison to other representative data science benchmarks is illustrated in Table~\ref{tab:comparison}.

\section{\datasetname: Benchmark data and tasks}

\datasetname is constructed from scientific publications and their associated biomedical data. At the core of the benchmark are structured components that mirror the research process: a curated collection of publications and corresponding data tables, extracted hypotheses paired with supporting evidence, and data analysis tasks derived from these elements. This framework supports the development and assessment of AI agents on a wide spectrum of capabilities, from code generation and reasoning to hypothesis testing, grounded in scientific discovery workflows. In the following subsections, we detail the construction of \datasetname, including how publications and data were collected, how hypotheses and supporting evidence were extracted, and how downstream tasks were defined to challenge and benchmark agent performance.

\begin{figure}[t]
    \centering
    \includegraphics[width=0.95\linewidth]{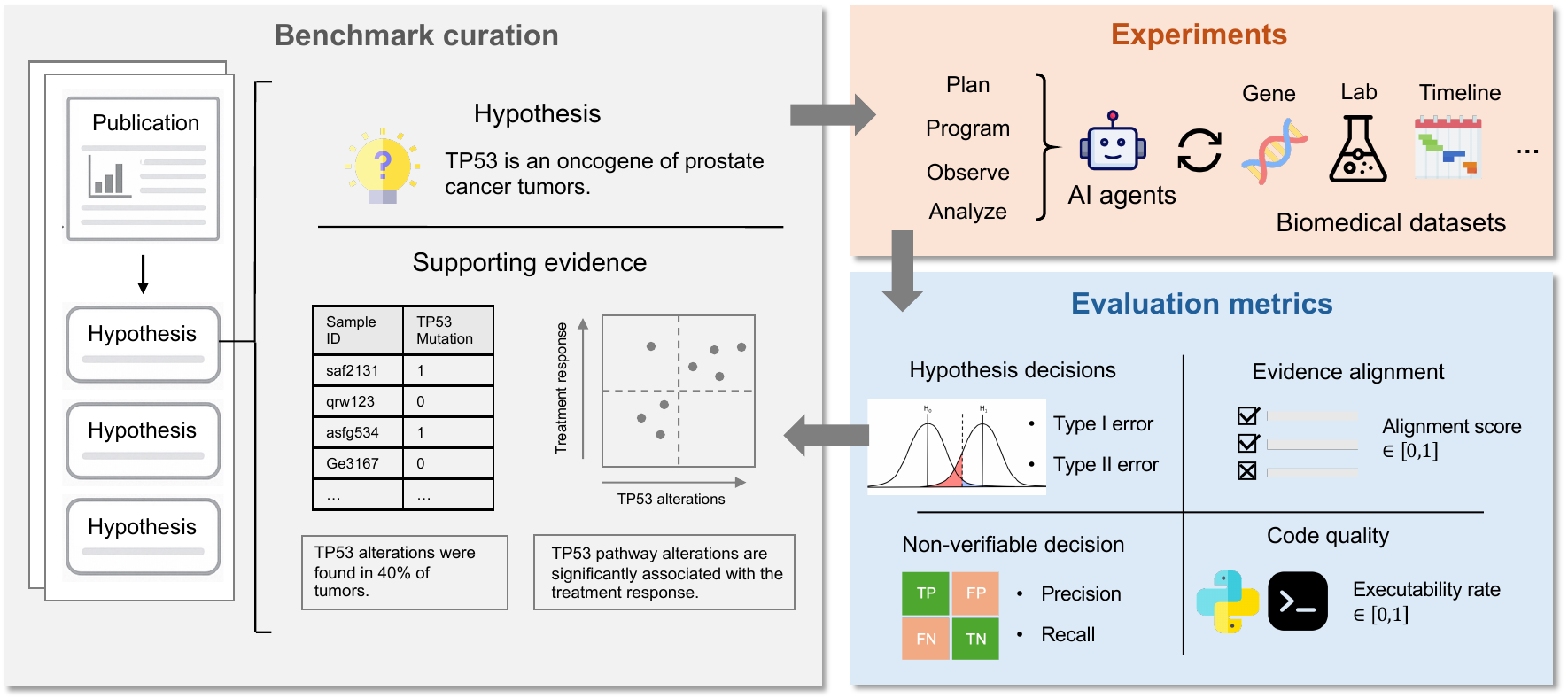}
    \caption{Overview of \datasetname. \textbf{a}, Benchmark curation: Scientific publications linked to biomedical datasets are parsed to extract hypotheses and their corresponding supporting evidence, forming the core reasoning challenges.
\textbf{b}, Experiments: AI agents are tasked with validating hypotheses by planning analysis steps, generating executable code, observing results, and making decisions based on structured biomedical datasets.
\textbf{c} Evaluation metrics: Agent performance is evaluated based on hypothesis decision accuracy (Type I and Type II errors), evidence alignment with publication findings, non-verifiable hypothesis detection (precision and recall), and code executability rate.}
    \label{fig:framework}
    \vspace{-1em}
\end{figure}

\subsection{Publication and dataset collection}
To construct a benchmark that reflects practical biomedical data science, it is essential to include not only scientific publications but also the corresponding biomedical datasets on which those studies are based. We therefore leverage cBioPortal~\cite{cbioportal2013}, a comprehensive cancer genomics and clinical data portal that maintains structured datasets with direct linkage to peer-reviewed publications. It is under a publicly available Open Database License~\cite{odbl}. This ensures that our benchmark captures both the analytical context and the quantitative evidence underlying published findings. In particular, we assume that each publication highlights its primary results within the abstract, often supported by descriptive statistics, statistical testing, and predictive modeling results derived from the associated biomedical tables. Thus, the raw data underlying \datasetname consists of two components: the publication abstracts and their corresponding structured biomedical data tables.

We utilize the cBioPortal API\footnote{\url{https://github.com/cbioportal/cbioportal/}} to retrieve all available datasets in bulk. Each dataset includes study metadata that specifies the associated publication(s), including PubMed identifiers (PMIDs). Using these PMIDs, we collect the publication abstracts through the PubMed API\footnote{\url{https://www.ncbi.nlm.nih.gov/home/develop/api/}}. In most cases, there is a one-to-one mapping between a dataset and a publication. However, we exclude ambiguous cases involving multiple papers and datasets when their analytical scope extends beyond the specific dataset. This filtering step avoids introducing non-verifiable hypotheses into the benchmark, thereby maintaining a clear linkage between reported findings and the underlying data.

According to established biomedical literature~\cite{vogelstein2013cancer}, we categorize the publications in our benchmark by study types. Definitions of these categories are provided in Appendix~\ref{appx:pub_type}. As illustrated in Figure~\ref{fig:data_stats_combined}, \datasetname spans a diverse array of study types, including genomics, integrative, therapeutics, biomarkers, translational, and molecular studies, along with various analysis methodologies. This distribution highlights the comprehensiveness of \datasetname, capturing both high-level exploratory research and focused hypothesis-driven studies.

\subsection{Dataset caption}
We caption the data tables for benchmarking in data science tasks while preserving privacy. The details of how the captioning works can be found in Appendix~\ref{appx:dataset_caption}. Specifically, we do not send any patient-level records to LLMs and instead construct a schema-based representation. For each column in a data table, we compute type-specific descriptive statistics, such as the number of unique values, missing value ratio, most frequent entries, and data ranges. In this way, for whatever LLM API provider we use, only the captions of the dataset will be shared. For future research and experiments with this benchmark, researchers can download the raw data from cBioPortal and execute the LLM-generated code on them locally.

Figure~\ref{fig:data_stats_combined} shows the scale and diversity of the biomedical tables included in \datasetname. Each point represents a data type, positioned by its typical number of rows and columns, and sized by its prevalence in the dataset. The benchmark encompasses a wide spectrum of commonly used biomedical data types, including clinical data, mutation data, gene expression, copy number alteration, protein expression, structural variation, and patient timelines. These data sources are foundational to modern biomedical research and collectively capture the heterogeneity of real-world biomedical analysis. Moreover, the wide variance in both row and column dimensions, ranging from compact gene panels to large-scale expression matrices, demonstrates the high dimensionality and analytical complexity present in \datasetname. Compared to existing benchmarks (as shown in Table~\ref{tab:comparison}), which often involve simpler, smaller, or less diverse datasets, our benchmark presents a significantly more challenging and realistic setting for evaluating AI agents on biomedical data science tasks.

\subsection{Hypothesis and evidence}
All data science challenges in \datasetname are extracted from published biomedical studies using a GPT-4o model. The details of the extraction process can be found in Appendix~\ref{appx:hypothesis_extraction}. Each challenge is centered around a hypothesis and its corresponding supporting evidence, reflecting how scientific claims are typically articulated in real-world literature. Rather than stating hypotheses solely in null form (e.g., ``no difference between groups''), authors of original studies often present claims affirmatively (e.g., ``Treatment A improves survival''), while the underlying analyses are grounded in statistical tests against a null hypothesis. To preserve fidelity to real-world practice, our benchmark follows this formulation, presenting hypotheses as definitive statements derived from the study's conclusions. Importantly, our design does not assume these statements are inherently true; instead, we evaluate whether AI agents can reconstruct the reasoning and analysis pipeline leading to such claims, including identifying when the data are insufficient to support them.

Each entry includes (1) a clearly stated \textit{hypothesis} that is supported or rejected in the original publication, and (2) a plausible \textit{counter-hypothesis} designed to test the agent’s ability to reason discriminatively. An example is provided in Supplementary Figure~\ref{fig:appx_hypothesis_example}. To support hypothesis validation, we extract one or more \textit{evidence} entries per hypothesis, each corresponding to a distinct data analysis performed in the study. Each evidence entry is annotated with the following fields:

\begin{itemize}[leftmargin=*]
    \item \textbf{Analysis plan:} a concise description of the statistical or computational procedure used (e.g., frequency analysis, correlation test, clustering).
    \item \textbf{Evidence:} a textual summary of the result as reported in the publication.
    \item \textbf{Variables:} input variables used in the analysis and the result variable serving as the output to support or refute the hypothesis.
\end{itemize}

\begin{table}[t]
\caption{Comparison of \datasetname with representative benchmarks in general and biomedical domains. ``Avg. \# Tables'' denotes the average number of tables per task; ``Avg. \# Columns'' refers to the average columns per table. ``–'' indicates missing or non-tabular data. ``\# Tasks'' shows the number of unique data science tasks. ``*'' indicates the biology-related portions of the benchmarks.}
\label{tab:comparison}
\resizebox{\textwidth}{!}{%
\begin{tabular}{lllllll}
\toprule
\textbf{Benchmark} & \textbf{Domain} & \textbf{Task Levels} & \textbf{Task Sources} & \textbf{Avg. \#  Tables} & \textbf{Avg. \# Columns*} & \textbf{\# Tasks} \\
\midrule
DS-1000~\cite{lai2023ds}             & General    & Analysis                & Stackoverflow       & 1 & -   & 1000 \\
MLAgentBench~\cite{huangmlagentbench}        & General    & Analysis                & Publications        & 1 & 47  & 13   \\
DSBench~\cite{jing2025dsbench}             & General    & Analysis                & Kaggle & - & -   & 466  \\
BLADE~\cite{gu2024blade}               & General    & Hypothesis and analysis & 31 Publications     & 1 & 13  & 12   \\
ScienceAgentBench~\cite{chen2024scienceagentbench}   & General    & Hypothesis and analysis & 44 Publications     & - & -   & 102  \\
\midrule
DiscoveryBench-Bio*~\cite{majumderdiscoverybench} & Biology    & Hypothesis and analysis & 2 Publications      & 2 & 26  & 16   \\
SciCode-Bio*~\cite{tian2024scicode}        & Biology    & Hypothesis and analysis & 8 Publications      & - & -   & 8    \\
BioCoder~\cite{tang2024biocoder}            & Biomedical & Analysis                & Github              & - & -   & 460  \\
ChatGPT-ADA~\cite{tayebi2024large}         & Biomedical & Hypothesis and analysis & 4 Publications      & 1 & 548 & 4    \\
AI Co-scientist~\cite{gottweis2025towards}     & Biomedical & Hypothesis and analysis & -                   & - & -   & 3    \\
BioDiscoveryAgent~\cite{roohani2024biodiscoveryagent}   & Biomedical & Analysis                & Publications        & 1 & -   & 6    \\
BioDSBench~\cite{wang2024can}          & Biomedical & Analysis                & 39 Publications     & - & -   & 293  \\
\midrule
\rowcolor{Azure2} \datasetname (ours)     & Biomedical & Hypothesis and analysis & 328 Publications    & 6 & 879 & 1029 \\
\bottomrule
\end{tabular}%
}
\end{table}

To mitigate bias toward Type I error, our benchmark includes a significant fraction of \textit{non-verifiable} cases where the available data are insufficient to reach a definitive conclusion. This design encourages agents not merely to ``prove'' hypotheses, but to assess them critically in the context of the available evidence, akin to a real-world research setting.

\subsection{Tasks and evaluation}
The primary task in \datasetname is hypothesis validation using structured biomedical data. Given a hypothesis extracted from a publication and the corresponding dataset, an AI agent is required to generate executable code to analyze the data and produce empirical observations. Based on these observations, the agent must decide whether the hypothesis is \textit{True}, \textit{False}, or \textit{Non-verifiable}. To distinguish between the latter two, we define a hypothesis as \textit{False} if the agent can identify relevant variables in the dataset and derive contradicting evidence through analysis. Conversely, a hypothesis is considered \textit{Non-verifiable} if no relevant features or data tables exist in the dataset to support or reject the claim. For example, the hypothesis ``Prostate cancer brain metastases (PCBM) have a higher mutational burden compared to non-brain metastases'' is labeled as \textit{Non-verifiable} if the dataset lacks mutational burden variables or comparative group labels.

We evaluate agent performance across multiple dimensions. On the hypothesis decision level, we compute both Type I and Type II error rates. Let $H \in {\text{True}, \text{False}}$ denote the ground truth label of a hypothesis, and $\hat{H}$ be the label predicted by the agent. The Type I error (false positive rate) is defined as:

\begin{equation} \text{Type I Error} = \frac{\sum \mathbb{I}[H = \text{False} \land \hat{H} = \text{True}]}{\sum \mathbb{I}[H = \text{False}]}, \end{equation}

and the Type II error (false negative rate) is given by:

\begin{equation} \text{Type II Error} = \frac{\sum \mathbb{I}[H = \text{True} \land \hat{H} = \text{False}]}{\sum \mathbb{I}[H = \text{True}]}, \end{equation}

where $\mathbb{I}[\cdot]$ is the indicator function.

In addition to correctness at the decision level, we assess how well the generated observations align with the supporting evidence reported in the original publication. Let $E$ denote the set of ground truth supporting evidences and $O$ the set of observations generated by the agent. We use a large language model (LLM)-as-a-judge~\cite{fu2024gptscore} approach to measure the evidence alignment score:

\begin{equation} \text{Alignment Score} = \frac{|O \cap E|}{|E|}. \end{equation}

This metric quantifies the proportion of reported evidence that is successfully captured by the agent's analysis pipeline.

Furthermore, we evaluate the technical quality of the generated code. For each hypothesis, let $C$ denote the total number of code cells generated and $C_{\text{exec}}$ the number of those that are executable without error. The code executability rate is defined as: $\text{Executability Rate} = \frac{C_{\text{exec}}}{C}$.

For ReAct-style agents that explore through multi-step reasoning, this metric is computed over all code snippets generated during the interaction trace.

Lastly, we systematically assess agents on their ability to reject non-verifiable hypotheses. These hypotheses are curated by taking claims from other publications that reference unrelated datasets. An ideal agent should classify such hypotheses as \textit{Non-verifiable} due to the absence of relevant data. Let $H = \text{Non-verifiable}$ be the ground truth and $\hat{H}$ be the predicted label. We report the non-verifiable detection accuracy as:

\begin{equation} \text{Non-verifiable Accuracy} = \frac{\sum \mathbb{I}[H = \text{Non-verifiable} \land \hat{H} = \text{Non-verifiable}]}{\sum \mathbb{I}[H = \text{Non-verifiable}]}. \end{equation}

This provides insight into how well agents can discern dataset limitations and avoid over-assertive conclusions.

\section{Experiment}

\subsection{Implemented methods}

\begin{wrapfigure}{r}{0.5\linewidth}
  \centering
  \includegraphics[width=\linewidth]{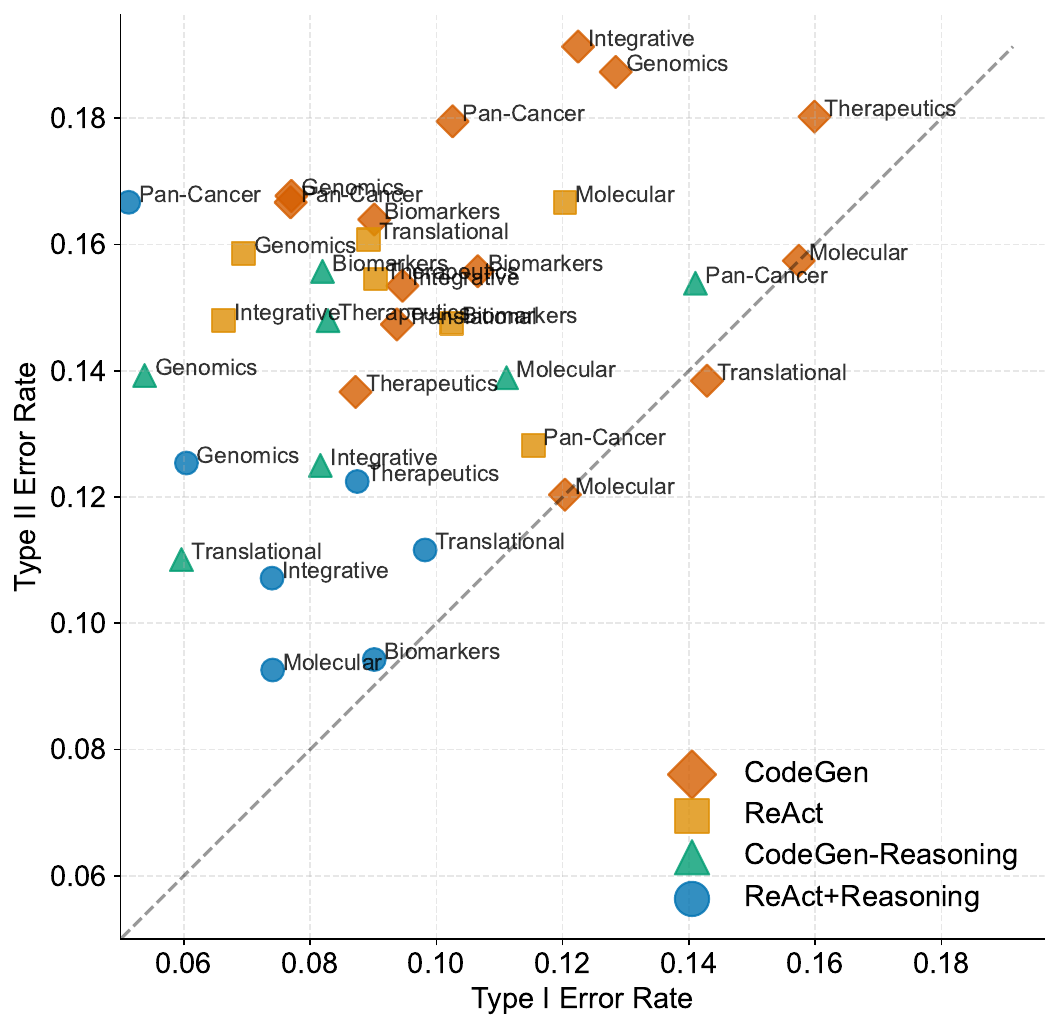}
  \caption{Comparison of Type I and Type II error rates across publication types and agent variants. Each point denotes an agent's performance on a specific publication type. 
}
  \label{fig:agent_pubtype_breakdown}
  \vspace{-1em}
\end{wrapfigure}

We implement four agent-based methods to evaluate performance on \datasetname. \textit{CodeGen} directly generates a single executable Python code block based on the input hypothesis and dataset schema, and returns a final decision, True, False, or Non-verifiable, based on the produced observations, without explicit intermediate reasoning~\cite{ridnik2024code}. We evaluate two variants of CodeGen: one powered by GPT-4o and the other by O3-mini. \textit{ReAct} follows the ReAct framework~\cite{yao2023react}, in which the agent alternates between reasoning steps (``thoughts'') and code execution (``actions''), allowing iterative refinement of analysis and conclusions. This version is implemented using GPT-4o.

To enable more structured reasoning, we also introduce two reasoning-augmented agents aligned with recent developments in data analysis agents~\cite{majumder2024position,huang2025automated}, which decouple experiment planning from execution. \textit{CodeGen-Reasoning} first prompts O3-mini to generate a structured analysis plan detailing key reasoning and statistical steps, and then passes this plan to GPT-4o for code generation and execution, allowing division of labor between planning and implementation. \textit{ReAct-Reasoning} extends ReAct with structured planning and uses O3-mini as the backend agent. It supports iterative reasoning and dynamic plan refinement based on intermediate observations across multiple steps.

\begin{table}[t]
  \centering
\caption{
Performance of hypothesis validation across publication types. 
Each cell reports the Type I error rate ($\err_I$, false positive rate) and Type II error rate ($\err_{II}$, false negative rate), with lower values indicating better performance. ``R*'' is short for ``Reasoning'' version of CodeGen and ReAct, respectively. Bold values highlight the best performance in each column.
}  \resizebox{\textwidth}{!}{
    \begin{tabular}{l|cc|cc|cc|cc|cc|cc|cc}
    \toprule
    \multicolumn{1}{c|}{\multirow{2}[2]{*}{Methods}} 
      & \multicolumn{2}{c|}{\makecell{Biomarkers\\(n=244)}} 
      & \multicolumn{2}{c|}{\makecell{Genomics\\(n=662)}} 
      & \multicolumn{2}{c|}{\makecell{Integrative\\(n=392)}} 
      & \multicolumn{2}{c|}{\makecell{Molecular\\(n=108)}} 
      & \multicolumn{2}{c|}{\makecell{Pan-Cancer\\(n=78)}} 
      & \multicolumn{2}{c|}{\makecell{Therapeutics\\(n=344)}} 
      & \multicolumn{2}{c}{\makecell{Translational\\(n=224)}} \\
          & $\err_I$ & $\err_{II}$ & $\err_I$ & $\err_{II}$ & $\err_I$ & $\err_{II}$ 
          & $\err_I$ & $\err_{II}$ & $\err_I$ & $\err_{II}$ & $\err_I$ & $\err_{II}$ 
          & $\err_I$ & $\err_{II}$ \\
    \midrule
    CodeGen (gpt-4o) & 0.090 & 0.164 & 0.077 & 0.168 & 0.095 & 0.153 & 0.157 & 0.157 & 0.077 & 0.167 & 0.087 & 0.137 & 0.094 & 0.147 \\
    CodeGen (o3-mini) & 0.107 & 0.145 & 0.128 & 0.187 & 0.122 & 0.191 & 0.098 & 0.118 & 0.103 & 0.179 & 0.157 & 0.181 & 0.143 & 0.138 \\
    ReAct (gpt-4o) & 0.102 & 0.148 & 0.069 & 0.159 & \textbf{0.066} & 0.148 & 0.120 & 0.167 & 0.115 & \textbf{0.128} & 0.090 & 0.155 & 0.089 & 0.161 \\
    CodeGen-R* & \textbf{0.082} & 0.156 & \textbf{0.054} & 0.139 & 0.082 & 0.125 & 0.111 & 0.139 & 0.141 & 0.154 & \textbf{0.083} & 0.148 & \textbf{0.060} & \textbf{0.110} \\
    ReAct-R* & 0.090 & \textbf{0.094} & 0.060 & \textbf{0.125} & 0.074 & \textbf{0.107} & \textbf{0.074} & \textbf{0.093} & \textbf{0.051} & 0.167 & 0.087 & \textbf{0.122} & 0.098 & 0.112 \\
    \bottomrule
    \end{tabular}%
    }
  \label{tab:hypothesis_validation}%
\end{table}%

\subsection{Hypothesis validation}
Table~\ref{tab:hypothesis_validation} shows that AI agents tend to be conservative in hypothesis validation across all tested publication types. In nearly every setting, the Type II error rate ($\err_{II}$), which measures the frequency of missed relevant findings, is consistently higher than the Type I error rate ($\err_{I}$), which reflects the incidence of false positives. For example, in the Biomarkers category, CodeGen (gpt-4o) exhibits a Type II error of 0.164 compared to a Type I error of 0.090.

Figure~\ref{fig:agent_pubtype_breakdown} and Table~\ref{tab:hypothesis_validation} also demonstrate that reasoning augmentation improves both sensitivity and specificity. Reasoning-enhanced agents (denoted with an asterisk, e.g., CodeGen-R* and ReAct-R*) consistently outperform their base counterparts in terms of lower error rates. For instance, ReAct-R* reduces the Type I and II errors in the Genomics category to 0.060 and 0.125, respectively, compared to 0.069 and 0.159 for the base ReAct model. Similarly, CodeGen-R* achieves a Type I error of 0.082 and Type II error of 0.156 on Biomarkers, outperforming the original CodeGen (gpt-4o) with errors of 0.090 and 0.164. These results indicate that structured reasoning enhances the agent’s ability to identify relevant evidence while reducing false positives.

\begin{wrapfigure}{r}{0.5\linewidth}
  \centering
  \includegraphics[width=\linewidth]{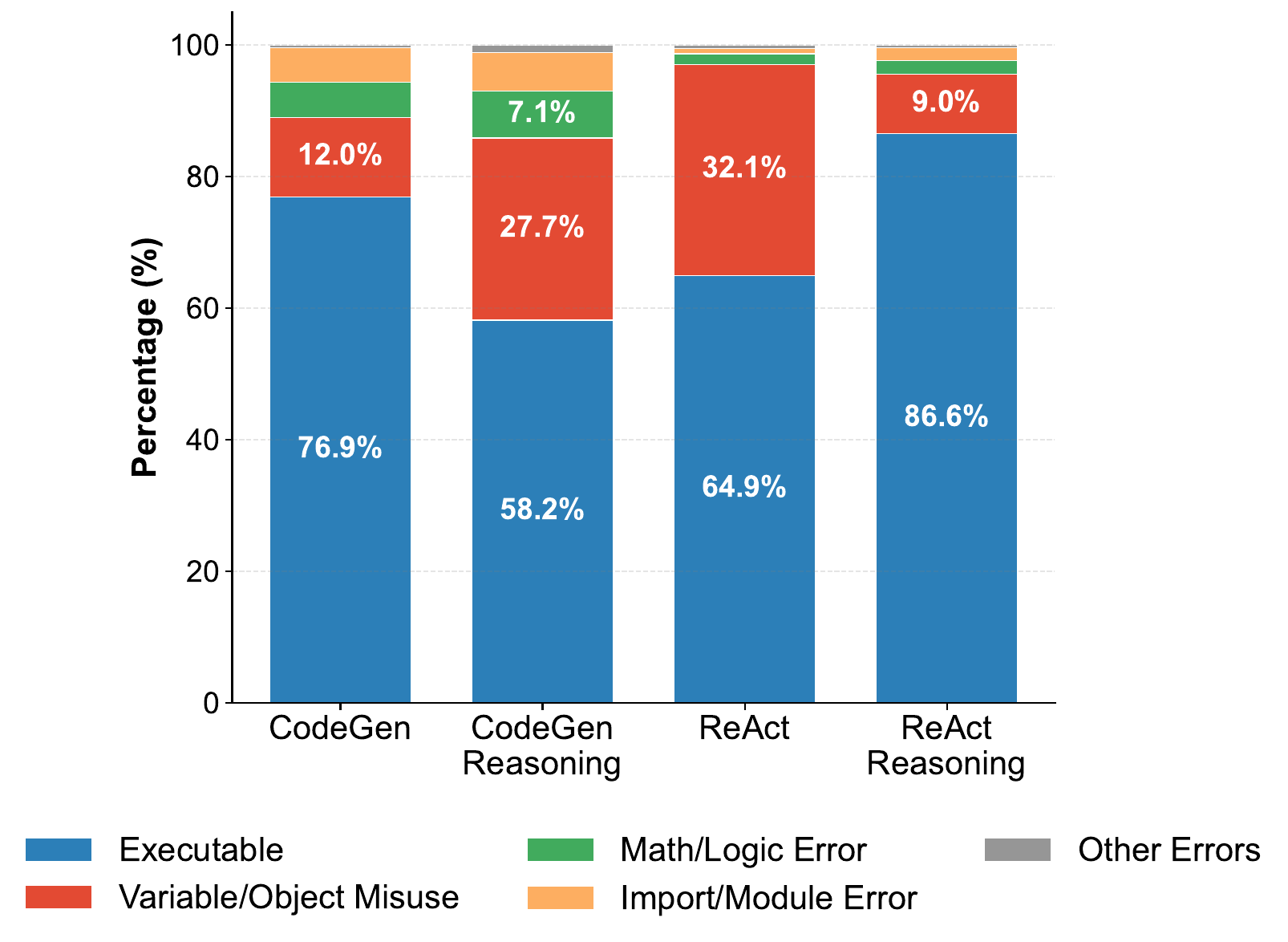}
  \caption{Code excitability analysis and the breakdown of error types in non-executable code across the selected AI agents.
}
  \label{fig:code_executability}
  \vspace{-1em}
\end{wrapfigure}

Figure~\ref{fig:agent_pubtype_breakdown} shows that ReAct-based methods consistently outperform CodeGen models, particularly when reasoning is applied. Even without reasoning, ReAct (gpt-4o) achieves lower Type II errors in challenging categories such as Integrative (0.148 vs. 0.153) and Pan-Cancer (0.128 vs. 0.167) compared to CodeGen (gpt-4o). When reasoning is incorporated, ReAct-R* outperforms CodeGen-R* in most domains, for example, in Translational, ReAct-R* reports a Type II error of 0.112 compared to 0.110 for CodeGen-R*, while maintaining a lower Type I error (0.098 vs. 0.060). 

Finally, Figure~\ref{fig:agent_pubtype_breakdown} suggests that reasoning brings the greatest improvements in domains with higher baseline error rates. This trend is evident in the Genomics and Integrative categories, where non-reasoning methods exhibit relatively high Type II errors: up to 0.191 for CodeGen (o3-mini) in Integrative. In contrast, ReAct-R* reduces the same error to 0.107. This implies that reasoning is particularly valuable in more complex or information-dense publication types, helping agents better navigate and resolve ambiguous or detailed hypotheses.

\begin{figure}[t]
    \centering
    \begin{minipage}{0.48\linewidth}
        \centering
        \includegraphics[width=\linewidth]{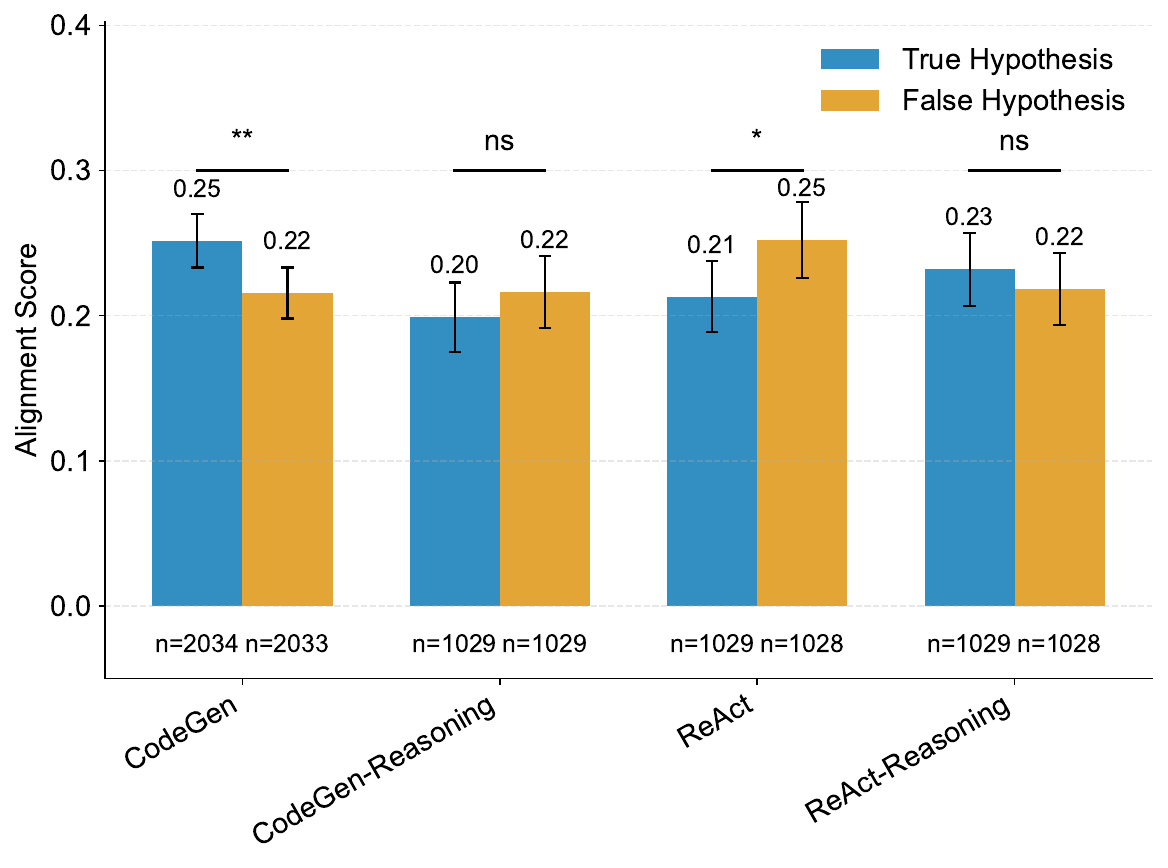}
        \caption{Evidence alignment scores for true vs. false hypotheses across methods.}
        \label{fig:figure4a}
    \end{minipage}%
    \hfill
    \begin{minipage}{0.48\linewidth}
        \centering
        \includegraphics[width=\linewidth]{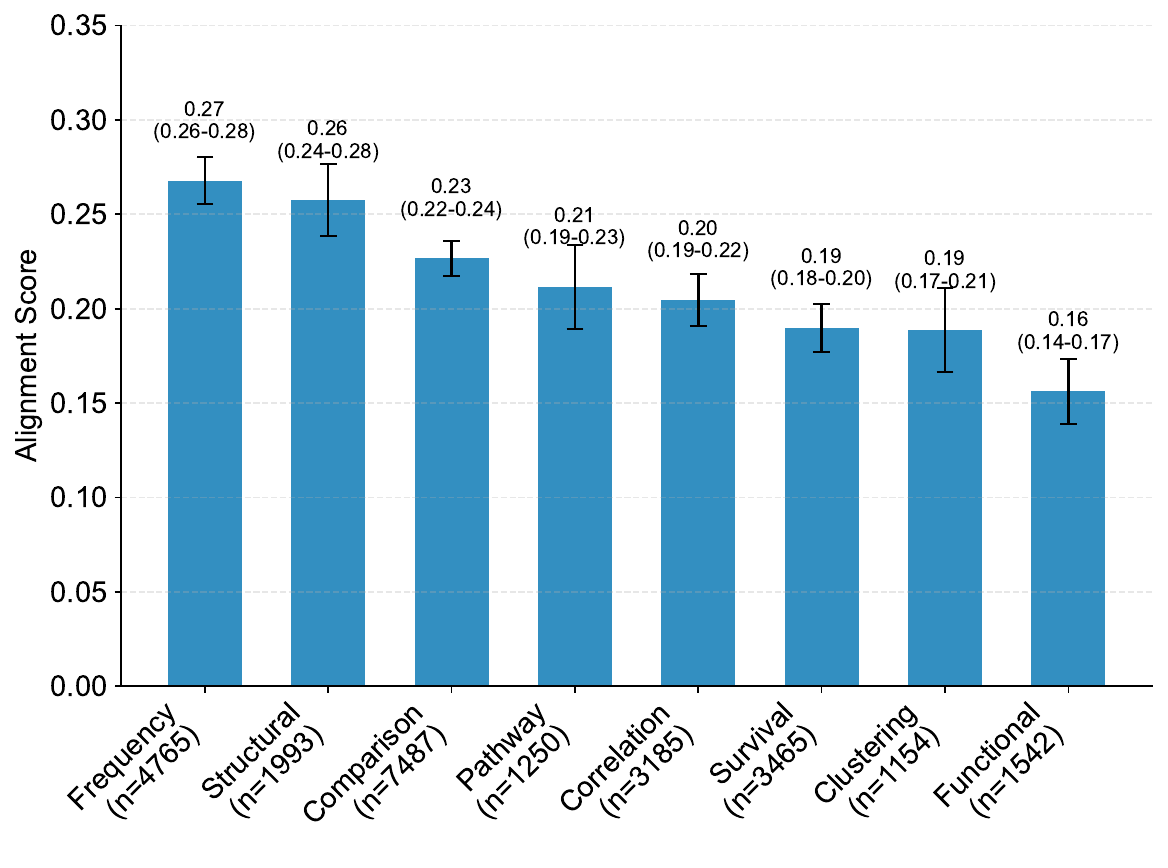}
        \caption{Evidence alignment scores by types of analyses.}
        \label{fig:figure4b}
    \end{minipage}
\end{figure}

\subsection{Analysis quality}
Making a correct hypothesis decision does not necessarily imply that the AI agent followed a valid or faithful analytical process, which is a limitation largely overlooked in prior evaluations. To address this, our benchmark explicitly assesses the evidence alignment score, which measures how well the agent-generated analysis captures the ground-truth evidence reported in the original studies. We also examine the executability of the analysis code produced by the agents as a proxy for code quality and practical usability.

As shown in Figure~\ref{fig:figure4a}, the evidence alignment scores remain modest across all methods, typically ranging from 0.20 to 0.25, regardless of whether the hypothesis being validated is ultimately True or False. Among the evaluated methods, ReAct-based agents exhibit marginally higher alignment scores compared to code generation baselines. However, the consistently low scores across the board suggest that AI agents often diverge from the evidence used in human-authored analyses, possibly reflecting a lack of domain knowledge or contextual understanding required for appropriate methodological choices.

Figure~\ref{fig:figure4b} further breaks down alignment scores by analysis type. We observe that simpler analytical tasks, such as frequency counts, are more reliably handled by AI agents. In contrast, more complex tasks, including clustering and survival analysis, pose significant challenges, with notably lower alignment scores across all models. These findings highlight the need for improved reasoning strategies and domain-specific modeling capabilities in AI systems aimed at biomedical data analysis.

Figure~\ref{fig:code_executability} presents the executability of the code generated by different AI agents and categorizes the types of errors found in non-executable outputs. Overall, ReAct-based agents exhibit the highest code executability rates, with ReAct Reasoning achieving 86.6\% and ReAct at 84.9\%, outperforming both CodeGen (76.9\%) and CodeGen Reasoning (58.2\%). Among the error types, variable or object misuse is the most common failure mode, especially prominent in CodeGen Reasoning (27.7\%) and ReAct (32.1\%). Logic and mathematical errors, as well as import or module-related issues, occur less frequently but still contribute to code failure across methods. 

\subsection{Non-verifiable hypothesis}

\begin{figure}
    \centering
    \includegraphics[width=0.8\linewidth]{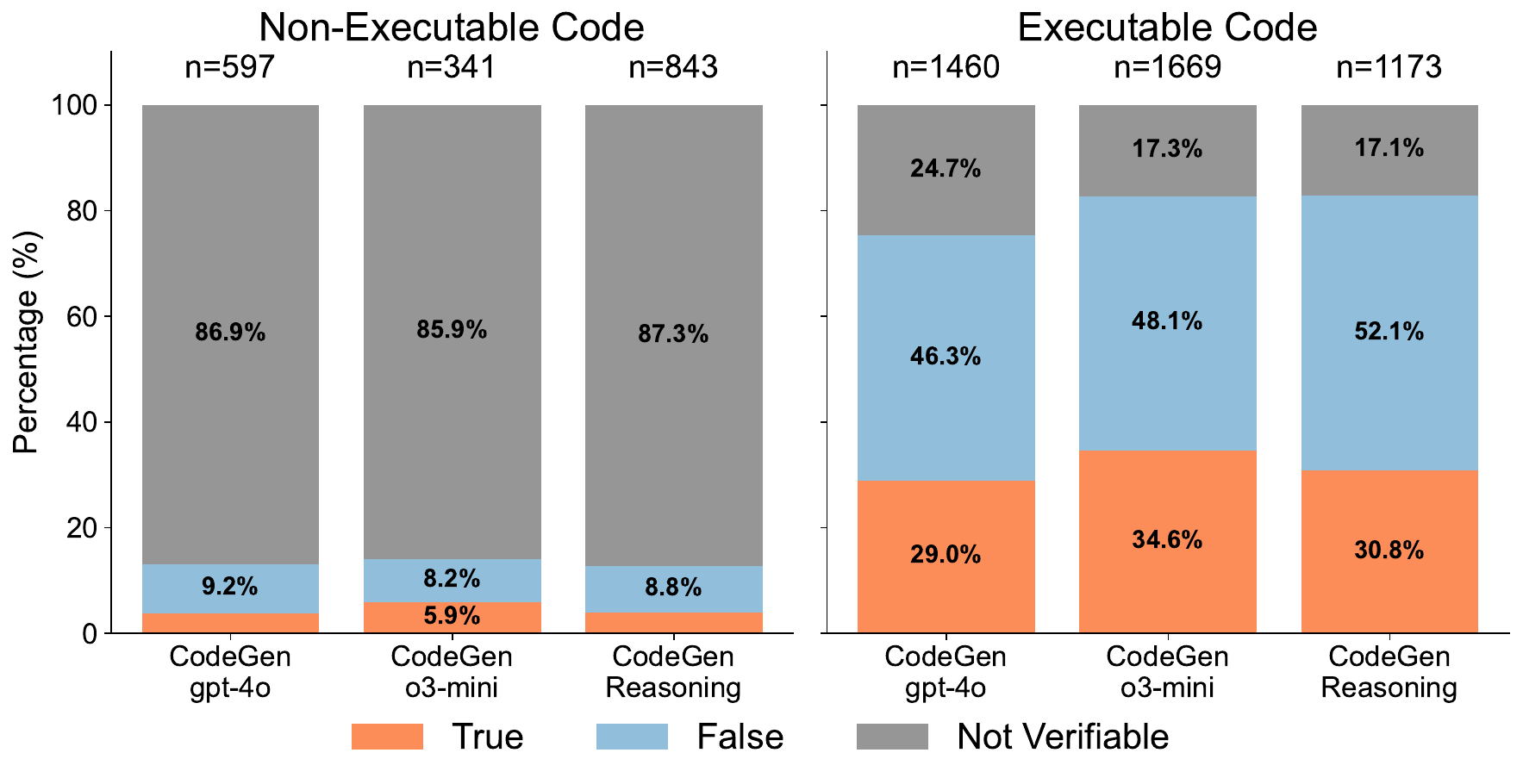}
    \caption{Hypothesis validation results' distribution by the code excitability for CodeGen methods.}
    
    \vspace{-1em}\label{fig:hypothesis_decision_non_executable_code}
\end{figure}

\begin{wrapfigure}{r}{0.5\linewidth}
  \centering
  \includegraphics[width=\linewidth]{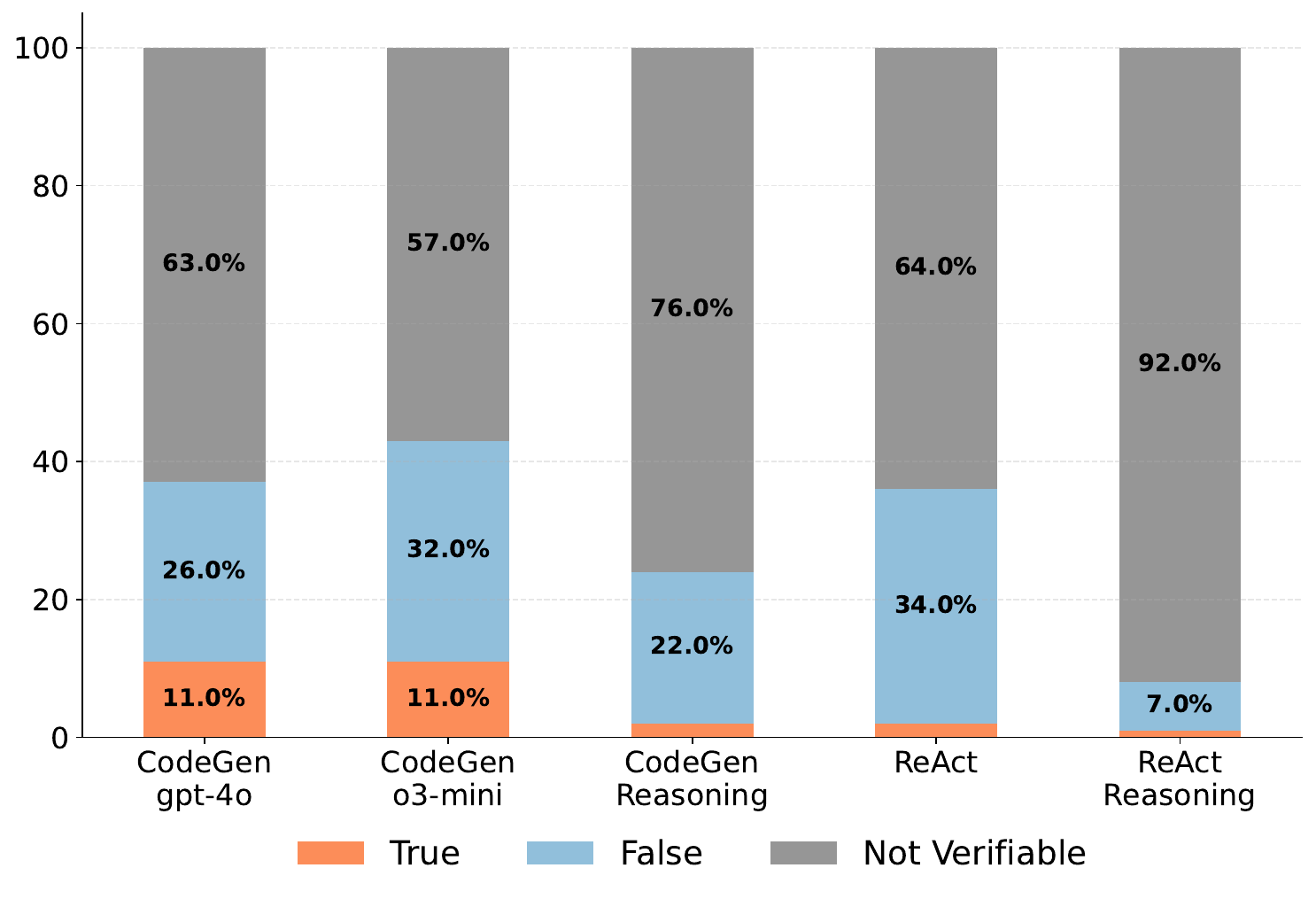}
  \caption{Hypothesis decision distribution for the non-verifiable hypothesis by different agents.
}
  \label{fig:nv_hypothesis}
  \vspace{-1em}    
\end{wrapfigure}

Figure~\ref{fig:code_executability} illustrates the proportion of executable versus non-executable code, stratified by error type, across different CodeGen methods. Since CodeGen and CodeGen-Reasoning only attempt a single-shot code generation, any failure in code execution should theoretically preclude meaningful hypothesis validation. To explore this, Figure~\ref{fig:hypothesis_decision_non_executable_code} presents the distribution of hypothesis decisions (True, False, Not Verifiable) based on whether the generated code was executable or not. The results reveal a marked difference in decision patterns between the executable and non-executable cases. In the non-executable setting, all three CodeGen variants default to deciding the hypothesis as Not Verifiable in approximately 87\% of instances, but still decide around 8\% as False and 5\% as True. These 13\% cases indicate that the AI agents sometimes turn out to hallucinate the findings. By contrast, in cases where the generated code is executable, the proportion of Not Verifiable decisions significantly drops, while the rates of True and False decisions increase substantially.

We further investigated whether AI agents can act cautiously when faced with non-verifiable hypotheses. As shown in Figure~\ref{fig:nv_hypothesis}, we constructed a set of 100 hypotheses that are strictly non-verifiable, meaning the associated dataset lacks the information needed to either accept or reject them. In this setting, the correct model behavior is to respond with ``Not Verifiable''; any decision of ``True'' or ``False'' reflects overconfidence or hallucination. The ability to correctly identify these cases, quantified as the true positive rate (TPR) for the ``Not Verifiable'' class, varies substantially across agents. One-round code generation methods, such as CodeGen gpt-4o and CodeGen o3-mini, achieve only 63\% and 57\% TPR, respectively, often making incorrect verifiable claims in 37\% and 43\% of the cases. In contrast, reasoning-augmented agents like CodeGen Reasoning, ReAct, and particularly ReAct Reasoning perform more conservatively, with ReAct Reasoning achieving a TPR of 92\%.

\section{Related work}
\paragraph{Benchmarks} Recent efforts have introduced benchmark datasets to evaluate AI agents in scientific discovery and data science tasks. General-purpose scientific discovery benchmarks such as DiscoveryBench~\citep{majumderdiscoverybench}, ScienceAgentBench~\citep{chen2024scienceagentbench}, and SpiderV2~\citep{cao2024spider2} focus on a broad range of tasks but often overlook specialized biomedical reasoning challenges. In parallel, several benchmarks specifically target the core task of code generation in scientific domains, including SciCode~\citep{tian2024scicode}, Blade~\citep{gu2024blade}, and DSBench~\citep{jing2025dsbench}. Within biomedicine, BioCoder~\citep{tang2024biocoder} and CliniDSBench~\citep{wang2024can} address coding tasks related to biomedical data analysis. However, these benchmarks primarily emphasize code generation, while our work focuses on the broader hypothesis validation process derived directly from published scientific studies. Moreover, \datasetname offers a significantly larger and more diverse evaluation scale, encompassing over three hundred publications, substantially exceeding the coverage of previous benchmarks.

\paragraph{Agents}
A growing body of work explores the use of AI agents for data science and scientific research. Several systems target general data science tasks, including machine learning modeling and analysis on structured datasets such as Kaggle competitions~\citep{guo2024ds,grosnit2024large,li2024autokaggle,huangmlagentbench}. \citet{gao2024empowering} emphasizes the potential of developing agents specifically tailored for biomedical research. In the biomedical domain, Co-Scientist~\citep{gottweis2025towards} and BioDiscoveryAgent~\citep{roohani2024biodiscoveryagent} focus on a niche area: automating the design and execution of genetic perturbation experiments. Other agent frameworks have applied LLMs for bioinformatics programming, biomedical question answering~\citep{mehandru2025bioagents}, and the development of predictive models for biological outcomes~\citep{tayebi2024large}. Closest to our work is the line of research on hypothesis validation agents~\citep{huang2025automated}, which investigates how agents can reason over structured data to accept or refute scientific claims. Our work builds on these foundations but uniquely grounds the validation tasks in hypotheses and evidence derived from real-world publications, enabling broader and more rigorous evaluation of biomedical data science agents.

\section{Discussion and conclusion}
While our benchmark draws from over 300 biomedical studies, it does not fully capture the diversity of the biomedical research landscape. The dataset naturally overrepresents well-established topics with high publication volume, potentially underrepresenting emerging areas or those with limited available data. This skew may influence model performance and generalizability, highlighting the need to continuously expand and rebalance the benchmark to reflect a wider spectrum of scientific inquiry. More broadly, as AI agents become increasingly capable of performing end-to-end data science tasks, they also introduce the risk of generating plausible but incorrect scientific claims. Without proper oversight, such systems could accelerate the propagation of false findings under the guise of data-driven analysis. Ensuring transparency, interpretability, and human-in-the-loop validation will be critical to responsibly deploying these tools in high-stakes scientific domains.

In this work, we present \datasetname{}, a benchmark designed to evaluate AI agents on realistic biomedical data science tasks. By extracting over a thousand hypotheses and corresponding analysis plans from hundreds of published studies, \datasetname{} captures the diversity and complexity inherent in real-world biomedical research. Unlike prior benchmarks, it encompasses not only hypothesis validation tasks with sufficient evidence, but also non-verifiable cases where the available data are inconclusive: a frequent yet underrepresented scenario in scientific reasoning. The benchmark enables comprehensive evaluation across multiple dimensions, including decision accuracy, evidence grounding, reasoning validity, and analysis code executability. We envision \datasetname{} as a foundation for developing more robust, transparent, and trustworthy AI agents for scientific discovery.

\bibliographystyle{unsrtnat}
\bibliography{main}

%%%%%%%%%%%%%%%%%%%%%%%%%%%%%%%%%%%%zzzzzz%%%%%%%%%%%%%%%%%%%%%%%%

\clearpage

\section*{NeurIPS Paper Checklist}

\begin{enumerate}

\item {\bf Claims}
    \item[] Question: Do the main claims made in the abstract and introduction accurately reflect the paper's contributions and scope?
    \item[] Answer: \answerYes{} % Replace by \answerYes{}, \answerNo{}, or \answerNA{}.
    \item[] Justification: \answerNA{}
    \item[] Guidelines:
    \begin{itemize}
        \item The answer NA means that the abstract and introduction do not include the claims made in the paper.
        \item The abstract and/or introduction should clearly state the claims made, including the contributions made in the paper and important assumptions and limitations. A No or NA answer to this question will not be perceived well by the reviewers. 
        \item The claims made should match theoretical and experimental results, and reflect how much the results can be expected to generalize to other settings. 
        \item It is fine to include aspirational goals as motivation as long as it is clear that these goals are not attained by the paper. 
    \end{itemize}

\item {\bf Limitations}
    \item[] Question: Does the paper discuss the limitations of the work performed by the authors?
    \item[] Answer: \answerYes{} % Replace by \answerYes{}, \answerNo{}, or \answerNA{}.
    \item[] Justification: \answerNA{}
    \item[] Guidelines:
    \begin{itemize}
        \item The answer NA means that the paper has no limitation while the answer No means that the paper has limitations, but those are not discussed in the paper. 
        \item The authors are encouraged to create a separate "Limitations" section in their paper.
        \item The paper should point out any strong assumptions and how robust the results are to violations of these assumptions (e.g., independence assumptions, noiseless settings, model well-specification, asymptotic approximations only holding locally). The authors should reflect on how these assumptions might be violated in practice and what the implications would be.
        \item The authors should reflect on the scope of the claims made, e.g., if the approach was only tested on a few datasets or with a few runs. In general, empirical results often depend on implicit assumptions, which should be articulated.
        \item The authors should reflect on the factors that influence the performance of the approach. For example, a facial recognition algorithm may perform poorly when image resolution is low or images are taken in low lighting. Or a speech-to-text system might not be used reliably to provide closed captions for online lectures because it fails to handle technical jargon.
        \item The authors should discuss the computational efficiency of the proposed algorithms and how they scale with dataset size.
        \item If applicable, the authors should discuss possible limitations of their approach to address problems of privacy and fairness.
        \item While the authors might fear that complete honesty about limitations might be used by reviewers as grounds for rejection, a worse outcome might be that reviewers discover limitations that aren't acknowledged in the paper. The authors should use their best judgment and recognize that individual actions in favor of transparency play an important role in developing norms that preserve the integrity of the community. Reviewers will be specifically instructed to not penalize honesty concerning limitations.
    \end{itemize}

\item {\bf Theory assumptions and proofs}
    \item[] Question: For each theoretical result, does the paper provide the full set of assumptions and a complete (and correct) proof?
    \item[] Answer: \answerNA{} % Replace by \answerYes{}, \answerNo{}, or \answerNA{}.
    \item[] Justification: This paper does not involve theoretical results.
    \item[] Guidelines:
    \begin{itemize}
        \item The answer NA means that the paper does not include theoretical results. 
        \item All the theorems, formulas, and proofs in the paper should be numbered and cross-referenced.
        \item All assumptions should be clearly stated or referenced in the statement of any theorems.
        \item The proofs can either appear in the main paper or the supplemental material, but if they appear in the supplemental material, the authors are encouraged to provide a short proof sketch to provide intuition. 
        \item Inversely, any informal proof provided in the core of the paper should be complemented by formal proofs provided in appendix or supplemental material.
        \item Theorems and Lemmas that the proof relies upon should be properly referenced. 
    \end{itemize}

    \item {\bf Experimental result reproducibility}
    \item[] Question: Does the paper fully disclose all the information needed to reproduce the main experimental results of the paper to the extent that it affects the main claims and/or conclusions of the paper (regardless of whether the code and data are provided or not)?
    \item[] Answer: \answerYes{} % Replace by \answerYes{}, \answerNo{}, or \answerNA{}.
    \item[] Justification: \answerNA{}
    \item[] Guidelines:
    \begin{itemize}
        \item The answer NA means that the paper does not include experiments.
        \item If the paper includes experiments, a No answer to this question will not be perceived well by the reviewers: Making the paper reproducible is important, regardless of whether the code and data are provided or not.
        \item If the contribution is a dataset and/or model, the authors should describe the steps taken to make their results reproducible or verifiable. 
        \item Depending on the contribution, reproducibility can be accomplished in various ways. For example, if the contribution is a novel architecture, describing the architecture fully might suffice, or if the contribution is a specific model and empirical evaluation, it may be necessary to either make it possible for others to replicate the model with the same dataset, or provide access to the model. In general. releasing code and data is often one good way to accomplish this, but reproducibility can also be provided via detailed instructions for how to replicate the results, access to a hosted model (e.g., in the case of a large language model), releasing of a model checkpoint, or other means that are appropriate to the research performed.
        \item While NeurIPS does not require releasing code, the conference does require all submissions to provide some reasonable avenue for reproducibility, which may depend on the nature of the contribution. For example
        \begin{enumerate}
            \item If the contribution is primarily a new algorithm, the paper should make it clear how to reproduce that algorithm.
            \item If the contribution is primarily a new model architecture, the paper should describe the architecture clearly and fully.
            \item If the contribution is a new model (e.g., a large language model), then there should either be a way to access this model for reproducing the results or a way to reproduce the model (e.g., with an open-source dataset or instructions for how to construct the dataset).
            \item We recognize that reproducibility may be tricky in some cases, in which case authors are welcome to describe the particular way they provide for reproducibility. In the case of closed-source models, it may be that access to the model is limited in some way (e.g., to registered users), but it should be possible for other researchers to have some path to reproducing or verifying the results.
        \end{enumerate}
    \end{itemize}

\item {\bf Open access to data and code}
    \item[] Question: Does the paper provide open access to the data and code, with sufficient instructions to faithfully reproduce the main experimental results, as described in supplemental material?
    \item[] Answer: \answerYes{} % Replace by \answerYes{}, \answerNo{}, or \answerNA{}.
    \item[] Justification: \answerNA{}
    \item[] Guidelines:
    \begin{itemize}
        \item The answer NA means that paper does not include experiments requiring code.
        \item Please see the NeurIPS code and data submission guidelines (\url{https://nips.cc/public/guides/CodeSubmissionPolicy}) for more details.
        \item While we encourage the release of code and data, we understand that this might not be possible, so “No” is an acceptable answer. Papers cannot be rejected simply for not including code, unless this is central to the contribution (e.g., for a new open-source benchmark).
        \item The instructions should contain the exact command and environment needed to run to reproduce the results. See the NeurIPS code and data submission guidelines (\url{https://nips.cc/public/guides/CodeSubmissionPolicy}) for more details.
        \item The authors should provide instructions on data access and preparation, including how to access the raw data, preprocessed data, intermediate data, and generated data, etc.
        \item The authors should provide scripts to reproduce all experimental results for the new proposed method and baselines. If only a subset of experiments are reproducible, they should state which ones are omitted from the script and why.
        \item At submission time, to preserve anonymity, the authors should release anonymized versions (if applicable).
        \item Providing as much information as possible in supplemental material (appended to the paper) is recommended, but including URLs to data and code is permitted.
    \end{itemize}

\item {\bf Experimental setting/details}
    \item[] Question: Does the paper specify all the training and test details (e.g., data splits, hyperparameters, how they were chosen, type of optimizer, etc.) necessary to understand the results?
    \item[] Answer: \answerYes{} % Replace by \answerYes{}, \answerNo{}, or \answerNA{}.
    \item[] Justification: \answerNA{}
    \item[] Guidelines:
    \begin{itemize}
        \item The answer NA means that the paper does not include experiments.
        \item The experimental setting should be presented in the core of the paper to a level of detail that is necessary to appreciate the results and make sense of them.
        \item The full details can be provided either with the code, in appendix, or as supplemental material.
    \end{itemize}

\item {\bf Experiment statistical significance}
    \item[] Question: Does the paper report error bars suitably and correctly defined or other appropriate information about the statistical significance of the experiments?
    \item[] Answer: \answerYes{} % Replace by \answerYes{}, \answerNo{}, or \answerNA{}.
    \item[] Justification: \answerNA{}
    \item[] Guidelines:
    \begin{itemize}
        \item The answer NA means that the paper does not include experiments.
        \item The authors should answer "Yes" if the results are accompanied by error bars, confidence intervals, or statistical significance tests, at least for the experiments that support the main claims of the paper.
        \item The factors of variability that the error bars are capturing should be clearly stated (for example, train/test split, initialization, random drawing of some parameter, or overall run with given experimental conditions).
        \item The method for calculating the error bars should be explained (closed form formula, call to a library function, bootstrap, etc.)
        \item The assumptions made should be given (e.g., Normally distributed errors).
        \item It should be clear whether the error bar is the standard deviation or the standard error of the mean.
        \item It is OK to report 1-sigma error bars, but one should state it. The authors should preferably report a 2-sigma error bar than state that they have a 96\% CI, if the hypothesis of Normality of errors is not verified.
        \item For asymmetric distributions, the authors should be careful not to show in tables or figures symmetric error bars that would yield results that are out of range (e.g. negative error rates).
        \item If error bars are reported in tables or plots, The authors should explain in the text how they were calculated and reference the corresponding figures or tables in the text.
    \end{itemize}

\item {\bf Experiments compute resources}
    \item[] Question: For each experiment, does the paper provide sufficient information on the computer resources (type of compute workers, memory, time of execution) needed to reproduce the experiments?
    \item[] Answer: \answerYes{} % Replace by \answerYes{}, \answerNo{}, or \answerNA{}.
    \item[] Justification: \answerNA{}
    \item[] Guidelines:
    \begin{itemize}
        \item The answer NA means that the paper does not include experiments.
        \item The paper should indicate the type of compute workers CPU or GPU, internal cluster, or cloud provider, including relevant memory and storage.
        \item The paper should provide the amount of compute required for each of the individual experimental runs as well as estimate the total compute. 
        \item The paper should disclose whether the full research project required more compute than the experiments reported in the paper (e.g., preliminary or failed experiments that didn't make it into the paper). 
    \end{itemize}
    
\item {\bf Code of ethics}
    \item[] Question: Does the research conducted in the paper conform, in every respect, with the NeurIPS Code of Ethics \url{https://neurips.cc/public/EthicsGuidelines}?
    \item[] Answer: \answerYes{} % Replace by \answerYes{}, \answerNo{}, or \answerNA{}.
    \item[] Justification: \answerNA{}
    \item[] Guidelines:
    \begin{itemize}
        \item The answer NA means that the authors have not reviewed the NeurIPS Code of Ethics.
        \item If the authors answer No, they should explain the special circumstances that require a deviation from the Code of Ethics.
        \item The authors should make sure to preserve anonymity (e.g., if there is a special consideration due to laws or regulations in their jurisdiction).
    \end{itemize}

\item {\bf Broader impacts}
    \item[] Question: Does the paper discuss both potential positive societal impacts and negative societal impacts of the work performed?
    \item[] Answer: \answerYes{} % Replace by \answerYes{}, \answerNo{}, or \answerNA{}.
    \item[] Justification: \answerNA{}
    \item[] Guidelines:
    \begin{itemize}
        \item The answer NA means that there is no societal impact of the work performed.
        \item If the authors answer NA or No, they should explain why their work has no societal impact or why the paper does not address societal impact.
        \item Examples of negative societal impacts include potential malicious or unintended uses (e.g., disinformation, generating fake profiles, surveillance), fairness considerations (e.g., deployment of technologies that could make decisions that unfairly impact specific groups), privacy considerations, and security considerations.
        \item The conference expects that many papers will be foundational research and not tied to particular applications, let alone deployments. However, if there is a direct path to any negative applications, the authors should point it out. For example, it is legitimate to point out that an improvement in the quality of generative models could be used to generate deepfakes for disinformation. On the other hand, it is not needed to point out that a generic algorithm for optimizing neural networks could enable people to train models that generate Deepfakes faster.
        \item The authors should consider possible harms that could arise when the technology is being used as intended and functioning correctly, harms that could arise when the technology is being used as intended but gives incorrect results, and harms following from (intentional or unintentional) misuse of the technology.
        \item If there are negative societal impacts, the authors could also discuss possible mitigation strategies (e.g., gated release of models, providing defenses in addition to attacks, mechanisms for monitoring misuse, mechanisms to monitor how a system learns from feedback over time, improving the efficiency and accessibility of ML).
    \end{itemize}
    
\item {\bf Safeguards}
    \item[] Question: Does the paper describe safeguards that have been put in place for responsible release of data or models that have a high risk for misuse (e.g., pretrained language models, image generators, or scraped datasets)?
    \item[] Answer: \answerYes{} % Replace by \answerYes{}, \answerNo{}, or \answerNA{}.
    \item[] Justification: \answerNA{}
    \item[] Guidelines:
    \begin{itemize}
        \item The answer NA means that the paper poses no such risks.
        \item Released models that have a high risk for misuse or dual-use should be released with necessary safeguards to allow for controlled use of the model, for example by requiring that users adhere to usage guidelines or restrictions to access the model or implementing safety filters. 
        \item Datasets that have been scraped from the Internet could pose safety risks. The authors should describe how they avoided releasing unsafe images.
        \item We recognize that providing effective safeguards is challenging, and many papers do not require this, but we encourage authors to take this into account and make a best faith effort.
    \end{itemize}

\item {\bf Licenses for existing assets}
    \item[] Question: Are the creators or original owners of assets (e.g., code, data, models), used in the paper, properly credited and are the license and terms of use explicitly mentioned and properly respected?
    \item[] Answer: \answerYes{} % Replace by \answerYes{}, \answerNo{}, or \answerNA{}.
    \item[] Justification: \answerNA{}
    \item[] Guidelines:
    \begin{itemize}
        \item The answer NA means that the paper does not use existing assets.
        \item The authors should cite the original paper that produced the code package or dataset.
        \item The authors should state which version of the asset is used and, if possible, include a URL.
        \item The name of the license (e.g., CC-BY 4.0) should be included for each asset.
        \item For scraped data from a particular source (e.g., website), the copyright and terms of service of that source should be provided.
        \item If assets are released, the license, copyright information, and terms of use in the package should be provided. For popular datasets, \url{paperswithcode.com/datasets} has curated licenses for some datasets. Their licensing guide can help determine the license of a dataset.
        \item For existing datasets that are re-packaged, both the original license and the license of the derived asset (if it has changed) should be provided.
        \item If this information is not available online, the authors are encouraged to reach out to the asset's creators.
    \end{itemize}

\item {\bf New assets}
    \item[] Question: Are new assets introduced in the paper well documented and is the documentation provided alongside the assets?
    \item[] Answer: \answerYes{} % Replace by \answerYes{}, \answerNo{}, or \answerNA{}.
    \item[] Justification: \answerNA{}
    \item[] Guidelines:
    \begin{itemize}
        \item The answer NA means that the paper does not release new assets.
        \item Researchers should communicate the details of the dataset/code/model as part of their submissions via structured templates. This includes details about training, license, limitations, etc. 
        \item The paper should discuss whether and how consent was obtained from people whose asset is used.
        \item At submission time, remember to anonymize your assets (if applicable). You can either create an anonymized URL or include an anonymized zip file.
    \end{itemize}

\item {\bf Crowdsourcing and research with human subjects}
    \item[] Question: For crowdsourcing experiments and research with human subjects, does the paper include the full text of instructions given to participants and screenshots, if applicable, as well as details about compensation (if any)? 
    \item[] Answer: \answerNA{} % Replace by \answerYes{}, \answerNo{}, or \answerNA{}.
    \item[] Justification: \answerNA{}
    \item[] Guidelines:
    \begin{itemize}
        \item The answer NA means that the paper does not involve crowdsourcing nor research with human subjects.
        \item Including this information in the supplemental material is fine, but if the main contribution of the paper involves human subjects, then as much detail as possible should be included in the main paper. 
        \item According to the NeurIPS Code of Ethics, workers involved in data collection, curation, or other labor should be paid at least the minimum wage in the country of the data collector. 
    \end{itemize}

\item {\bf Institutional review board (IRB) approvals or equivalent for research with human subjects}
    \item[] Question: Does the paper describe potential risks incurred by study participants, whether such risks were disclosed to the subjects, and whether Institutional Review Board (IRB) approvals (or an equivalent approval/review based on the requirements of your country or institution) were obtained?
    \item[] Answer: \answerNA{} % Replace by \answerYes{}, \answerNo{}, or \answerNA{}.
    \item[] Justification: \answerNA{}
    \item[] Guidelines:
    \begin{itemize}
        \item The answer NA means that the paper does not involve crowdsourcing nor research with human subjects.
        \item Depending on the country in which research is conducted, IRB approval (or equivalent) may be required for any human subjects research. If you obtained IRB approval, you should clearly state this in the paper. 
        \item We recognize that the procedures for this may vary significantly between institutions and locations, and we expect authors to adhere to the NeurIPS Code of Ethics and the guidelines for their institution. 
        \item For initial submissions, do not include any information that would break anonymity (if applicable), such as the institution conducting the review.
    \end{itemize}

\item {\bf Declaration of LLM usage}
    \item[] Question: Does the paper describe the usage of LLMs if it is an important, original, or non-standard component of the core methods in this research? Note that if the LLM is used only for writing, editing, or formatting purposes and does not impact the core methodology, scientific rigorousness, or originality of the research, declaration is not required.
    %this research? 
    \item[] Answer: \answerNA{}{} % Replace by \answerYes{}, \answerNo{}, or \answerNA{}.
    \item[] Justification: \answerNA{}{}
    \item[] Guidelines:
    \begin{itemize}
        \item The answer NA means that the core method development in this research does not involve LLMs as any important, original, or non-standard components.
        \item Please refer to our LLM policy (\url{https://neurips.cc/Conferences/2025/LLM}) for what should or should not be described.
    \end{itemize}

\end{enumerate}

\appendix

\newpage
\DoToC
\newpage

\captionsetup[table]{name=Supplementary Table}
\setcounter{figure}{0}
\renewcommand*{\figurename}{Supplementary Figure}

\section{Example hypothesis and supporting evidence}

\begin{figure}[h]
    \centering
    \includegraphics[width=0.98\linewidth]{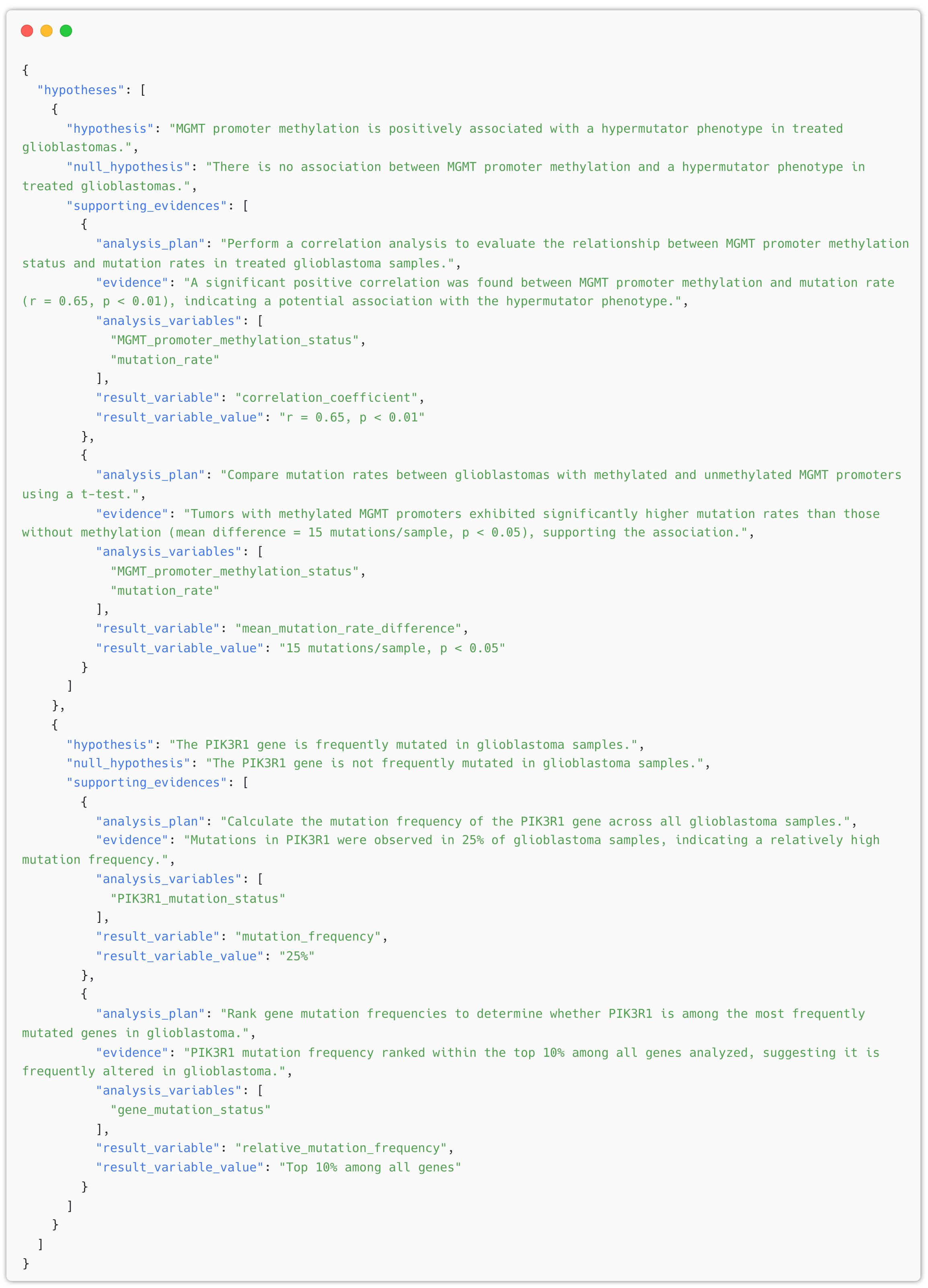}
    \caption{Examples of the hypothesis, counter-hypothesis, and supporting evidence extracted from biomedical publications.}
    \label{fig:appx_hypothesis_example}
\end{figure}

To illustrate the structure of entries in \datasetname{}, we present two representative examples derived from glioblastoma studies. Each example consists of a hypothesis formulated from the original study’s conclusions, a corresponding null hypothesis, and multiple supporting analyses that provide evidence for or against the claim.

\textbf{Example 1: MGMT Methylation and Hypermutation.}  
The hypothesis states that MGMT promoter methylation is positively associated with a hypermutator phenotype in treated glioblastomas. The corresponding null hypothesis asserts no such association. In one analysis, the study performed a correlation analysis between MGMT promoter methylation status and mutation rates, reporting a statistically significant positive correlation ($r = 0.65$, $p < 0.01$), suggesting that higher methylation is linked to increased mutation burden. A second analysis compared mutation rates between tumors with methylated versus unmethylated MGMT promoters, showing that methylated tumors had significantly higher mutation rates (mean difference = 15 mutations per sample, $p < 0.05$). These analyses collectively support the hypothesis.

\textbf{Example 2: PIK3R1 Mutation Frequency.}  
This hypothesis posits that the PIK3R1 gene is frequently mutated in glioblastoma samples, with the null hypothesis stating that it is not. In the first analysis, the study reported a mutation frequency of 25\% for PIK3R1 across glioblastoma samples, indicating a notable prevalence. A second analysis ranked gene mutation frequencies and found that PIK3R1 was among the top 10\% of all mutated genes, further supporting the claim of frequent alteration.

These examples demonstrate how \datasetname{} captures both the semantic structure of biomedical hypotheses and the analytical reasoning used to evaluate them, providing a grounded framework for assessing the capabilities of AI agents in data-driven scientific inference.

\section{Captions of Biomedical Data Tables}\label{appx:dataset_caption}

To support automated hypothesis validation and dataset reasoning tasks, we systematically generated structured captions for biomedical data tables from the cBioPortal repository. Each caption describes the content and structure of a tabular dataset, including its schema, value distributions, and metadata annotations. We developed a modular pipeline to process every dataset directory and extract metadata from text-based tables, primarily those with filenames beginning with \texttt{data\_}.

For each dataset, we extracted high-level metadata including dataset ID, cancer type, and description. Within each dataset directory, we identified all data tables and parsed their contents while ignoring comment lines (those beginning with ``\#''). The first non-comment line was interpreted as the column header. Subsequent lines were parsed as tab-delimited rows, with short rows padded and long rows truncated to maintain schema alignment.

To ensure consistency and facilitate downstream usage, we cleaned column names by removing punctuation and replacing whitespace with underscores. We then inferred the data type of each column using a custom heuristic function and computed column-wise statistics depending on the inferred type:
\begin{itemize}[leftmargin=*]
    \item \textbf{Binary and categorical columns:} Top value counts and number of unique values were reported, along with missing value rate.
    \item \textbf{Integer-valued columns:} We computed quantiles (1\%, 20\%, 40\%, 60\%, 80\%, 99\%) as well as minimum and maximum values.
    \item \textbf{Continuous columns:} Descriptive statistics were generated, including count, mean, standard deviation, and range, rounded to four decimal places.
\end{itemize}

Each table’s caption includes its name, number of rows and columns, column-level statistics, and preserved comment rows if present. The final structured metadata was saved as JSON files under a centralized metadata directory, one per dataset. These captions serve as machine-readable documentation for real-world biomedical tables and are critical for enabling dataset-aware reasoning by AI agents.

An example metadata structure is as follows:
\begin{verbatim}
{
  "dataset_id": "example_ds",
  "type_of_cancer": "glioblastoma",
  "description": "Clinical and genomic data for GBM samples",
  "tables": [
    {
      "name": "data_clinical.txt",
      "n_rows": 287,
      "n_columns": 12,
      "n_comment_rows": 4,
      "columns": [
        {
          "name": "age_at_diagnosis",
          "data_type": "integer",
          "n_unique": 45,
          "missing_rate": 0.03,
          "statistics": {
            "min": 22,
            "0.2": 45,
            "0.8": 73,
            "max": 88,
            "statistics_type": "quantiles"
          }
        },
        ...
      ]
    }
  ]
}
\end{verbatim}

This structured captioning process enables interpretability, reusability, and intelligent query capabilities across diverse biomedical datasets in \datasetname{}.

\section{Extracting hypothesis and evidence from publications}\label{appx:hypothesis_extraction}
To construct a benchmark of data-driven scientific claims, we developed a large-scale pipeline for extracting testable hypotheses and their supporting evidence from biomedical publications. Each extracted instance consists of a binary hypothesis derived from the abstract, a plausible counter-hypothesis, and one or more structured evidence entries grounded in quantitative findings.

We began with a curated metadata file from cBioPortal, which includes over 1,000 publications indexed by PubMed ID (PMID), their associated dataset identifiers, and accompanying titles, abstracts, and result summaries. After deduplication and filtering, we paired each publication with its corresponding abstract and dataset identifiers. For each entry, we concatenated the title and abstract to form a unified context and sent it to a large language model (GPT-4o) via a structured prompt.

The prompt was designed to elicit hypotheses that are:
\begin{itemize}[leftmargin=*]
    \item Binary and testable using statistical or machine learning methods,
    \item Grounded in measurable outcomes, with clear references to statistical relationships or effect sizes,
    \item Accompanied by structured supporting evidence, including analysis plans, involved variables, statistical measures, and result values.
\end{itemize}

\begin{lstlisting}
# The prompt:
"""
The following is the abstract of a publication:

{abstract}

Task:
Given the abstract of a publication, your task is to extract binary hypotheses and their supporting evidences that can be tested through data analysis.

Requirements for hypotheses and evidences:
1. Each hypothesis must be testable using statistical analysis or machine learning methods
2. All evidence must include specific, measurable quantities or statistical relationships
3. Result values must be numerical (e.g., percentages, counts, p-values, correlation coefficients) or categorical with clear classifications
4. Analysis variables must be specific data columns or features that exist in the dataset

Return your answer as a JSON object in the following format:
```json
{
    "hypotheses": [
        {
            "hypothesis": a specific, binary hypothesis that can be tested statistically, from the abstract, the one which is considered to be true from the study,
            "wrong_hypothesis": make a random perturbation of the hypothesis so that it is a wrong hypothesis,
            "supporting_evidences": [  // the evidences that support the alternative hypothesis
                {
                    "analysis_plan": a brief analysis plan that can yield this evidence,
                    "evidence": specific statistical finding or measurement,
                    "analysis_variables": list of exact variables/features needed for analysis,
                    "result_variable": the specific metric or statistical measure used,
                    "result_variable_value": numerical value, statistical measure, or categorical outcome
                },
                ...
            ]
        },
        ...
    ]
}
"""
\end{lstlisting}

To increase throughput and reliability, we used batched LLM calls with zero temperature to ensure deterministic completions. Each LLM output was parsed using a custom function that handled both valid JSON and malformed output formats via regular expression matching. The output structure follows a fixed schema that includes a \texttt{hypothesis}, a \texttt{wrong\_hypothesis} (a small perturbation to simulate a plausible counterfactual), and a list of \texttt{supporting\_evidences}, each containing fields such as \texttt{analysis\_plan}, \texttt{evidence}, \texttt{analysis\_variables}, \texttt{result\_variable}, and \texttt{result\_variable\_value}.

Each extracted hypothesis is linked to its source publication (via PMID) and the relevant datasets (via dataset ID) so that the claim can later be validated against real-world biomedical tables. The full outputs were stored as structured JSON files, one per publication. This corpus forms the foundation of \datasetname{}, enabling AI agents to reason over realistic, evidence-backed scientific claims.

\section{Categorization of biomedical publications}\label{appx:pub_type}

\paragraph{Genomics}  
Publications in this category focus on large-scale genomic profiling of tumors. These studies utilize high-throughput sequencing to catalog somatic mutations, copy-number variations, and other genetic alterations across cancer samples.

\paragraph{Molecular}  
This class covers research that investigates molecular characteristics beyond DNA mutations. It includes analyses of transcriptomic, proteomic, and epigenomic data, often derived from both patient samples and established cancer cell lines.

\paragraph{Pan-Cancer}  
Pan-Cancer studies undertake comparative analyses across multiple types of cancers. They aim to identify common molecular patterns and differences, thereby deepening our understanding of shared and unique cancer pathways.

\paragraph{Therapeutics}  
These publications explore the relationship between genomic alterations and drug responses. The focus is on identifying potential therapeutic targets and advancing personalized treatment strategies based on genetic and molecular data.

\paragraph{Biomarkers}  
Research in this class is dedicated to discovering and validating diagnostic and prognostic markers. These biomarkers help in predicting disease outcomes, guiding treatment decisions, and supporting early detection.

\paragraph{Methods}  
Publications categorized as Methods introduce new computational tools, algorithms, or experimental techniques that facilitate the analysis and interpretation of complex biomedical data.

\paragraph{Integrative}  
Integrative studies combine data from multiple omics layers—such as genomics, transcriptomics, and proteomics—to provide a comprehensive view of tumor biology. They aim to interconnect disparate data types into coherent biological insights.

\paragraph{Translational:}  
This class emphasizes bridging the gap between research and clinical application. Translational studies apply genomic and molecular findings to improve diagnostic methods, prognostic assessments, and treatment strategies in clinical practice.

\section{Categorization of analysis tasks}\label{appx:analysis_type}

\paragraph{Correlation analysis (Correlation)} Tasks that focus on statistically relating two or more variables. Examples include correlating gene methylation status with mutation rates or associating mutational profiles with clinical factors such as smoking status.

\paragraph{Comparative analysis (Comparison)} Tasks that directly contrast groups or conditions. These include comparing mutation frequencies between groups (e.g., methylated vs. unmethylated promoters) or contrasting profiles across different cancer subtypes.

\paragraph{Frequency analysis (Frequency)} Tasks that measure the occurrence or rate of specific genomic events. Typical examples are calculating the mutation frequency for a given gene or determining the prevalence of a particular genetic alteration.

\paragraph{Clustering and classification (Clustering)} Tasks that involve grouping data based on similarities. These studies might use cluster analysis to categorize samples by genomic features or mutational signatures.

\paragraph{Survival and prognostic analysis (Survival)} Tasks that associate molecular or genomic features with patient outcomes, such as survival curve comparisons or prognostic evaluations.

\paragraph{Functional and experimental analysis (Functional)} Tasks that explore gene function or cellular behavior through experimental approaches. This includes RNA interference experiments or assays measuring the effects of gene knockdown on cell proliferation.

\paragraph{Genomic structural analysis (Structural)} Tasks that analyze genomic architecture or structural variants. Examples include the evaluation of copy-number alterations, genomic rearrangements, or spatial mutation distributions.

\paragraph{Pathway and integrative analysis (Pathway)} Tasks that integrate multiple data types to elucidate biological pathways and networks. These include integrative pathway analyses, enrichment studies, or assessments of driver mutations in signaling cascades.

\section{Categorization of code errors}
To enable systematic analysis of common failure modes in AI-generated code, we group low-level Python error types into broader categories reflecting common code quality issues. This many-to-one mapping provides a more interpretable summary of model behaviors and facilitates downstream visualization and comparison. The categorization is defined as follows:

\begin{itemize}[leftmargin=*]
    \item \textbf{Variable/Object Misuse:} Errors such as \texttt{KeyError}, \texttt{AttributeError}, \texttt{NameError}, and \texttt{IndexError} that arise from referencing undefined variables, missing dictionary keys, or invalid object attributes.
    
    \item \textbf{Math/Logic Error:} Includes errors like \texttt{ZeroDivisionError}, \texttt{ValueError}, and \texttt{numpy.linalg.LinAlgError}, which typically result from invalid arithmetic operations, numerical instability, or logical violations.
    
    \item \textbf{Import/Module Error:} Consists of \texttt{ImportError} and \texttt{ModuleNotFoundError}, indicating missing dependencies or incorrect import paths.
    
    \item \textbf{File/I-O Error:} Captures input/output-related issues such as \texttt{FileNotFoundError} and \texttt{OSError}, often caused by referencing unavailable files or malformed I/O operations.
    
    \item \textbf{Pandas/Data Error:} Includes errors from data processing libraries, such as \texttt{pandas.errors.ParserError}, \texttt{MergeError}, and \texttt{IndexingError}, typically caused by invalid parsing, merging, or indexing operations.
    
    \item \textbf{General Exception:} Encompasses generic or runtime-specific errors such as \texttt{Exception} and \texttt{RuntimeError}, which represent critical failures not captured by more specific categories.
\end{itemize}

\section{Agent prompts}
\subsection{CodeGen}
\begin{lstlisting}
# Prompts for CodeGen methods:

"""
# TASK
Given the user-provided scientific hypothesis, you **Must** write {language} code to help the user evaluate the hypothesis.

# IMPORTANT: CODE OUTPUT REQUIREMENTS
You must import all the necessary libraries at the beginning of your code.

You must use explicit print() statements for ALL outputs you want to see or analyze. Simply writing expressions like 'df.head()' will NOT show results in the execution log. Always use:
- print(df.head())
- print(analysis_result)
- print(statistical_test_output)
Every intermediate result and final output must be wrapped in a print() statement to be visible in the execution log.


# DATASET PATHS
{dataset_paths}

# DATASET SCHEMA  
{dataset_schema}

## Ouptut
Your output should be in Markdown format and you should wrap the generated code in ```{language} ``` tags.
"""
\end{lstlisting}

\subsection{ReAct}
\begin{lstlisting}
# Prompts for ReAct methods:

"""
# TASK
Evaluate the user's scientific hypothesis using the datasets provided. You can write and execute {language} code to evaluate the hypothesis by invoking the tool {tool_name}.

# IMPORTANT: CODE OUTPUT REQUIREMENTS
You must import all the necessary libraries at the beginning of your code.

You must use explicit print() statements for ALL outputs you want to see or analyze. Simply writing expressions like 'df.head()' will NOT show results in the execution log. Always use:
- print(df.head())
- print(analysis_result)
- print(statistical_test_output)
Every intermediate result and final output must be wrapped in a print() statement to be visible in the execution log.

# DATASET SCHEMA
{dataset_schema}

# DATASET PATHS
{dataset_paths}
"""
\end{lstlisting}

\subsection{CodeGen-Reasoning}

\begin{lstlisting}
# CodeGen-Reasoning prompts

ANALYSIS_PLAN_PROMPT_TEMPLATE = """
# TASK  
Generate an analysis plan to evaluate the user's scientific hypothesis using the datasets provided.

The plan should consist of clear, actionable steps that can be **easily converted to {language} code** without needing any additional information.

# REQUIREMENTS  
- Use only table and column names from the schema: do not invent or guess names.  
- Ensure every step is unambiguous and directly executable.  
- Use consistent naming for all variables (e.g., tables, columns) throughout the plan.  
- Be as concise as possible while maintaining full clarity and precision.

# DATASET PATHS  
{dataset_paths}

# DATASET SCHEMA  
{dataset_schema}

# OUTPUT FORMAT  
Wrap the analysis plan in <analysis_plan> </analysis_plan> tags. So an example output would be
```
<analysis_plan>
1. load the dataset
2. print hello world
</analysis_plan>
"""

CODE_GENERATION_PROMPT_TEMPLATE = """
# TASK
Given the user-provided analysis plan for the user's scientific hypothesis, you **Must** write {language} code to fulfill the plan so that user can execute the code later
to evaluate the hypothesis.

# IMPORTANT: CODE OUTPUT REQUIREMENTS
You must import all the necessary libraries at the beginning of your code.

You must use explicit print() statements for ALL outputs you want to see or analyze. Simply writing expressions like 'df.head()' will NOT show results in the execution log. Always use:
- print(df.head())
- print(analysis_result)
- print(statistical_test_output)
Every intermediate result and final output must be wrapped in a print() statement to be visible in the execution log.


# DATASET PATHS
{dataset_paths}

## Ouptut
Your output should be in Markdown format and you should wrap the generated code in ```{language} ``` tags.
"""
\end{lstlisting}

\subsection{ReAct-Reasoning}
\begin{lstlisting}
AGENT_MODEL_PROMPT_TEMPLATE = """
You are a scientific agent who can plan and execute python code iteratively to evaluate a scientific hypothesis.

Note:
- You must execute and refine the given analysis plan iteratively until you have enough evidence to support the hypothesis.
- You must always write a single Python code block that can be executed directly based on the analysis plan.
- Use `print()` statements in your code to get the observations.
"""

PLANNING_PROMPT_TEMPLATE = """
# TASK  
Generate an analysis plan to evaluate the user's scientific hypothesis using the datasets provided.

The plan should consist of clear, actionable psudo codesteps that can be **easily converted to python code** without needing any additional information.

# REQUIREMENTS 
- Use only table and column names from the schema: do not invent or guess names.  
- Ensure every step is unambiguous and directly executable.  
- Use consistent naming for all variables (e.g., tables, columns) throughout the plan.  
- Be as concise as possible while maintaining full clarity and precision.

# DATASET PATHS
{dataset_paths}

# DATASET SCHEMA
{dataset_schema}
"""
\end{lstlisting}

%%%%%%%%%%%%%%%%%%%%%%%%%%%%%%%%%%%%%%%%%%%%%%%%%%%%%%%%%%%%

\end{document}